\documentclass[a4paper]{article}
\usepackage{aaai}
\usepackage{times}
\usepackage{helvet}
\usepackage{courier}
\nocopyright
\usepackage[english]{babel}
\newcommand{\st}{\medskip\noindent}
\newcommand{\stn}{\medskip}
\def\ni{\noindent}

\def\beq{\begin{equation}}
\def\eeq#1{\label{#1}\end{equation}}
\long\def\COMMENT#1\ENDCOMMENT{\message{(Commented text...)}\par}

\def\and{ \ \wedge}

\def\ar{\leftarrow}
\newcommand{\rar}{\rightarrow}
\newcommand{\then}{\Rightarrow}

\newcommand{\nin}{\not\in}

\newcommand{\no}{\hbox{\it not}\ }

\newcommand{\cali}{{\cal I}}

\newcommand{\bbbb}{{\rm I\!B}}

\newtheorem{definition}{Definition}
\newtheorem{proposition}{Proposition}
\newtheorem{theorem}{Theorem}
\newtheorem{example}{Example}

\begin{document}

\title{Characterizing and computing stable models of logic programs:\\
the non--stratified case}

\author{G. Brignoli\\
Universit\`{a} di Milano\\
Dip.\ di Scienze dell'Informazione \\
I-20135 Milano, Italy\\
{\it brigno@ntboss.tesi.dsi.unimi.it}\\
\And S. Costantini\\
Universit\`{a} di L'Aquila\\
Dip.\ di Matematica Pura e Applicata \\
I-67100 L'Aquila, Italy\\
{\it stefcost@univaq.it}
\And O. D'Antona \and A. Provetti\\
Universit\`{a} di Milano\\
Dip.\ di Scienze dell'Informazione \\
I-20135 Milano, Italy\\
{\it \{dantona,provetti\}@dsi.unimi.it}
}

\maketitle

Stable Logic Programming (SLP) is an emergent, alternative style of logic programming:
 each solution to a problem is represented by a stable model of a deductive
database/function--free logic program encoding the problem itself.
Several implementations now exist for stable logic programming, and their 
performance is rapidly improving.
To make SLP generally applicable, it should be possible to check for
consistency (i.e., existence of stable models) of the input program before
attempting to answer queries.
In the literature, only rather strong sufficient conditions have been proposed 
for consistency, e.g., stratification.
This paper extends these results in several directions. 
First, the syntactic features of programs, viz. cyclic negative dependencies,
affecting the existence of stable models are characterized, and their relevance is discussed.
Next, a new graph representation of logic programs, 
the Extended Dependency Graph {\it (EDG),} is
introduced, which conveys enough information for reasoning about stable models
(while the traditional Dependency Graph does not).
Finally, we show that
the problem of the existence of stable models can be reformulated in terms of
 coloring of the {\it EDG}.

\COMMENT
\section{Introduction} 
This paper addresses the problem of the existence of stable models for deductive
databases and function--free logic programs and, more generally, of computing stable models.
There are pragmatic reasons for the great attention that stable models
now receive in knowledge representation and deductive databases.
In fact, stable models enable us to represent default assumptions, constraints, 
uncertainty and nondeterminism {\em in a direct way} 
(see \cite{BarGel94} for a discussion), and are a bridge between logic
programming and nonmonotonic reasoning theory. 
The automation of nonmonotonic reasoning may well rely upon automatic stable
model computation, through the well-studied equivalences with, for
  instance, autoepistemic logic \cite{MarTru91}.
In a recent paper \cite{MarTru99}, Marek and Truszczy\`nski
 have proposed the stable model
semantics as a foundation for a new programming style, called
Stable Models Programming (SLP), which departs significantly from conventional, 
Prolog-based logic programming. 

Many implemented systems are now being developed for computing stable models,
among them {\sc smodels} \cite{NieSim98} \cite{Sim97}, {\sc DeReS} \cite{ChoMarTru96},
{\sc SLG} \cite{CheWar96}, and {\sc dlv} \cite{Eit+97}.
 
Thanks, (and not despite) these recent significant progresses, 
we believe that more work is needed,
namely towards  establishing significant syntactic and semantic conditions for
  the existence of stable models.
Results are so far available only for the subclass of stratified logic programs
\cite{AptBol94} and for call--consistent programs \cite{Dun92}.
On the other hand, examples concerning non--stratified programs in literature
mostly regards programs such as planners that have models {\em by construction.}
Still, non-stratified programs capture aspects of commonsense reasoning 
that would not be easily representable otherwise, 
such as incomplete knowledge about the initial situation of a planning
domain or nondeterministic choice of an element 
from a set in deductive databases \cite{SacZan97}. 
This is the reason for the recent attention in literature for the problem of
computing stable models of programs in the general case.

The main contributions of this paper are:
(i) the definition of a set of syntactic and semantic conditions on
programs, that characterize the existence of stable models, and
(ii) the formulation of these conditions in terms of
a new class of graph associated to a program, leading to simple definitions. 

Precisely, the first contribution consists in observing that the 
syntactic structure of the program is a key factor affecting the existence and 
the number of stable models, while the standard notion of 
Dependency graph of a program is unable to capture this aspect.

The second contribution consists in 
introducing the {\em Extended dependency graph} (EDG) representation as
an isomorphism from programs to directed graphs.
Thence, we are able to characterize stable models in terms of {\em
admissible} colorings of the EDG of the program.
\ENDCOMMENT

\section{Background definitions}\label{background}
The stable model semantics \cite{GelLif88} is a view of logic programs as sets
of inference 
rules (more precisely,  default inference rules), where a stable
model is a set of atoms closed under the program itself.
Alternatively, one can see a program as a set of constraints on the solution of
a problem, where each
stable model represents a solution compatible with the constraints expressed by the program.  
Consider the simple program $\{ q \ar \no p, \no c.\ \ p \ar \no q.\ p\ar c.\}$. 
For instance, the first {\em rule} is read as ``assuming  that both  $p$ and $c$ 
are false, we can {\em conclude} that $q$ is true.'' 
This program has two stable models. 
In the first, $q$ is true while $p$ and $c$ are false; in the second, $p$ is
true while $q$ and $c$ are false. 

Unlike with other semantics, a program may have no stable model,
i.e., be contradictory, like the following: $\{ a \ar \no b.\ b \ar \no c.\ c \ar
\no a. \}$, where no set of atoms is closed under the rules.
It is important to make sure that a program admits stable models before
attempting to perform deduction.
Inconsistency may arise, realistically, when programs are combined: 
if they share atoms, a
subprogram like that above may {\sl surface} in the resulting program.

In this paper we consider, essentially, the language $DATALOG^{\neg}$ for
deductive databases, which is more restricted than traditional logic programming.
As discussed in \cite{MarTru99}, this restriction is not a limitation at this stage.

A rule (clause) $\rho$ is defined as usual, and can be seen 
as composed of a conclusion $head(\rho)$, and 
a set of conditions $body(\rho)$, the latter divided into positive conditions 
$pos(\rho)$ and negative conditions $neg(\rho)$.

For syntax and semantics of logic
programs with negation ({\em general}, or {\em normal} logic programs), 
and for the definition of Dependency Graph (DG), the reader may refer
for instance to \cite{AptBol94} and to the references therein.

\ni
For the sake of clarity however, let us report the
definition of stable models. We start from 
the subclass of positive programs, 
i.e. those where, for every rule $\rho$, $neg(\rho) = \emptyset$.

\begin{definition} (Stable model of positive programs)

\ni
The {\em stable model} $a(\Pi)$ of a positive program $\Pi$ is the
smallest subset of $\bbbb_{\Pi}$ such that for any rule $a\ar a_1,\dots a_m$ in 
$\Pi$: $a_1,\ldots,a_m \in a(\Pi) \then a\in a(\Pi)$.

\end{definition}\label{posStableModel}

\ni
Positive programs are unambiguous, in that they
have a unique stable model, which coincides 
with that obtained applying other semantics.

\begin{definition} (Stable models of programs)\label{StableModel}

\ni
Let $\Pi$ be a logic program. For any set $S$ of atoms, let $\Pi^S$ be a 
program obtained from $\Pi$ by deleting 
(i) each rule that has a formula `$\no\ A$' in its body with $A \in S$, and 
(ii) all formulae of the form `$\no\ A$' in the bodies of the remaining rules.

$\Pi^S$ does not contain {\it ``not,''} so that its stable model is 
already
defined. If this stable model coincides with $S$, then we say that $S$ is a
{\em stable model} of $\Pi$.
In other words, the stable models of $\Pi$ are characterized by the equation: 
$S=a(\Pi^S).$
\end{definition}

\ni
Programs which have a unique stable model are called categorical. 

\ni
In the literature, the main (sufficient) condition to ensure 
the existence of stable models is
call--consistency, which is summarized as follows: no atom {\em depends} on
itself via an odd number of negative conditions. 

\begin{proposition} \cite{Dun92}\label{dungtheorem}
A [normal] logic program has a stable model if it is call--consistent.  
\end{proposition}

More results along the lines of Proposition \ref{dungtheorem} 
are found in \cite{Dun92}.
However, this condition is quite restrictive, as there are programs with 
{\em odd cycles} (in the
sense described above) that have one or more stable models. 
See Example \ref{samegraph1} below.

For the sake of simplicity, 
in this paper we consider {\em kernel programs,}
that are general logic programs where: (i) there are no positive conditions, 
i.e. for every clause $\rho$, $pos(\rho)=\emptyset$; (ii) every atom which
is the head of a rule must also appear in the body of some rule (possibly the same one).

From any program $\Pi$, a kernel program $ker(\Pi)$ can be obtained, which is equivalent 
to $\Pi$ as far as characterizing stable models.

\section{The relationship between cycles and stable models}\label{oldwork}
As discussed above, we are interested in programs that 
are not stratified ({\em unstratified} programs), and do not satisfy call-consistency.
We will speak of an {\em even} (resp. {\em odd}) cycle
referring to a even (resp. odd) number of rules organized like $\{a\ar \no b.\
b\ar \no a.\}$ (resp. $\{c\ar \no e.\ e\ar \no f.\ f\ar \no c.\}$).
This Section is devoted to the analysis of the type and number of 
cycles appearing in a program,
and their {\em connections},
i.e, roughly, rules involving atoms that
appear in different cycles, which we call {\em handles}.
We argue that the form of cycles and connections is {\em the key factor} 
affecting the existence --and the number-- of stable models.
In fact, the dependency graph makes neither the cycles, nor the connections explicit.

\begin{example} \label{samegraph1}

\ni
Consider the following programs, $\Pi_1$, $\Pi_2$ and $\Pi_3$:

\st\ni $
\begin{array}{lll}
p\ar \no p, \no e.       & p\ar \no p.                 & p\ar \no p, \no e. \\
a\ar \no b.              & p\ar\no e.                  & a\ar \no b. \\
b\ar \no a.              & a\ar \no b.                 & b\ar \no a.\\
e\ar \no f.              & b\ar \no a.                 & e\ar \no f.\\
f\ar \no h.              & e\ar \no f.                 & f\ar \no h. \\
h\ar \no e.              & f\ar \no h.                 & h\ar \no e, \no a. \\
h\ar \no a.              & h\ar \no e, \no a.          &
\end{array} $
\end{example}

\ni
It is easy to see that the dependency graphs of the three programs 
in Example \ref{samegraph1} coincide.
However, $\Pi_1$ has the stable model $\{b,h,e\}$ while instead $\Pi_2$
has the stable model $\{a,f,p\}$ and $\Pi_3$ has no stable models at all.
Why do they have such a diverse semantics?
The reason relies in the different decomposition of the three programs into
cycles.
The programs above are divided into cycles as follows,
where {\it OC} and {\it EC} denote odd and even cycle, respectively, and
literals appearing either in square brackets or in braces correspond to
different kinds of handles. 
Consider the following partitions of $\Pi_1$ and $\Pi_2$, respectively:
\renewcommand{\arraycolsep}{.03in}

\st\ni $
\begin{array}{rlrl}  
OC_1: & \left\{      
\begin{array}{l}        
p\ar \no p, [\no e].    
\end{array}             
\right. 
&  OC_1: & \left\{
\begin{array}{l}
 p\ar \no p.  
\end{array}
\right.\\
EC_1: & \left\{      
\begin{array}{l}      
a\ar \no b.\\          
b\ar \no a.           
\end{array}           
\right.
&  H_2: & \left\{ 
\begin{array}{l}
p\ar \{\no e.\}\\  
\end{array}
\right.\\
OC_2: & \left\{        
\begin{array}{l}                           
e\ar \no f.\\                      
f\ar \no h.\\                      
h\ar \no e.           
\end{array}         
\right.
&  EC_1: & \left\{
\begin{array}{l}
a\ar \no b.\\ 
b\ar \no a.
\end{array}
\right.\\          
H_1: & \left\{      
\begin{array}{l}                               
h\ar \{\no a\}.                 
\end{array}             
\right.
&  OC_2: & \left\{
\begin{array}{l}  
e\ar \no f.\\
f\ar \no h.\\   
h\ar \no e, [\no a].  
\end{array} 
\right.
\end{array}
$

\st
The literals in braces are called {\it OR handles} of the cycle. 
Consider program $\Pi_1$.
Literal $\no a$ in $H_1$ is an OR handle for $OC_2$. 
Now, consider a putative stable model $S$; if  $a\nin S$, we can say that
``handle $H_1$ is true.'' Then, atom $g$ is {\sl forced} to be in $S$ and,
consequently, $OC_2$ has, w.r.t. $S$, the stable model $\{g, e\}$.
Literal $\no e$, instead, is an {\it AND handle} (indicated in square brackets) of the odd cycle $OC_1$:
if it is false (i.e., $e \in S$), it forces $p$ to be false, and $OC_1$ ``has
the empty model,'' and $p\nin S$.

Similar considerations can be made on $\Pi_2$, even tough it has a different structure: literal $\no a$ in this case is an AND handle to $OC_2$ (while in
$\Pi_1$ it is an OR handle, instead); if $\no a$ is true then the odd
cycle $OC_2$ is contradictory, and determines the inconsistency 
of the whole program.
If, on the other hand, $\no a$ is false, then $g$ is forced to be false,
and consequently $OC_2$ has the stable model $\{f\}$. 
This means moreover that the OR handle $\no e$ of $OC_1$ is 
true, and thus $p$ is true: therefore the contradiction $p\ar \no p$,
which could determine the inconsistency 
of the whole program, is made harmless.
Finally, the reader can easily check that program $\Pi_3$ has the odd cycle 
$OC_2$ unconstrained (no handles); thus, $\Pi_3$ has no stable models.
A formal assessment of cycles will be part of the forthcoming extended 
version of this paper.

At this point, it is however important to notice that one rule may belong
 to several cycles at once.

\begin{example}\label{manycycles} 
Let $\Pi_4:$

\st $
\begin{array}{ll}
p \ar \no p,\no q.  & a \ar \no b.\\
q \ar \no q, \no p. & b \ar \no a.\\
q \ar \no v.        & z \ar \no z, \no k.\\
v \ar \no w.        & k \ar \no l.\\
w \ar \no a.        & l \ar \no k.
\end{array} $

\st
In $\Pi_4$, the following cycles are found: \\
$C_1= \{ p \ar \no p, \no q.\}$\\
This is an odd cycle ($p$ depends on itself).\\
$C_2= \{ q \ar \no q, \no p.\}$\\
This is an odd cycle ($q$ depends on itself). 
Moreover, the former two rules together
form also an even cycle, where $p$ depends on $q$ and vice versa, i.e.:\\
$C_3= \{ q \ar \no p, \no q. \ p \ar \no q, \no p.\}$\\
Now,\\
$C_4= \{ a \ar \no b.\ b \ar \no a \}$\\
is an even cycle, while \\
$C_5= \{ z \ar \no z, \no k.\}$\\
is an odd cycle ($z$ depends on itself). Finally, \\
$C_6= \{ k \ar \no l. \ l \ar \no k.\}$\\
is an even cycle, $k$ depends on $l$ and vice versa.
There are clauses, namely $q \ar \no v.$,
$v \ar \no w.$, and $w \ar \no a.$, which do not belong to any cycle. Notice
however that they can be seen as {\em forming a chain connecting cycles.}
In fact, since the fist atom in the chain is $q$,
which belongs to cycles $C_2$ and $C_3$,
 in a way this chain forms two {\em bridges:} one between $C_2$ and $C_4$,
a the other between $C_3$ and $C_4$.
\end{example}


In Example \ref{manycycles} above, clause $q \ar \no v.$ is called an {\em auxiliary} rule of
cycles $C_2$ and $C_3$, since its conclusion $q$ is an atom belonging 
to these cycles. As mentioned above, auxiliary rules can belong to a {\em bridge} connecting different cycles.
For the sake of simplicity, we can assume that all bridges have unitary
length, i.e. that all bridges reduce to an auxiliary rule. 
In fact, what is important is which cycle is connected to
which, while the intermediate steps of the chain do not affect
the existence and number of stable models. 

In the rest of the paper, we will say that a
cycle $C$ is  {\em constrained} if it has an handle.
Then, a cycle with no handle is called {\em unconstrained}.

\subsection{From cycles to stable models}\label{cycletosm}
In order to reason about the existence of the stable models 
of $\Pi$, it is useful to reason about the existence of
the stable models of {\em its composing cycles}.

\begin{definition}
An extended cycle $EC$ is a set of rules composed of one cycle $C$
together with all its auxiliary clauses.
\end{definition}

\begin{proposition}
A program $\Pi$ has a unique decomposition into extended cycles
 $\{EC_{1},\ldots,EC_{r}\}$, $r \geq 1$.
\end{proposition}

\begin{definition}
Let $C$ be an extended cycle, and let ${H_C}$ be the set of all the atoms occurring
in some of the handles of $C$.
Let $\cali \in 2^{H_C}$. A completed extended cycle $CC$ is a set of rules composed of one 
extended cycle $C$, where atoms in $\cali$ are added as unit clauses.
\end{definition}

Notice that adding to $EC$ some of the atoms of $H_C$ (which 
are atoms occurring in the handles of $C$) corresponds to 
{\em making an hypothesis} about truth/falsity of the handles of C.
For any extended cycle $EC$, there are $2^{H_C}$ corresponding
completed cycles, each one corresponding to a different hypothesis on
the handles.
Correspondingly, there are several ways of decomposing 
$\Pi$ into completed cycles
 $\{CC_{1},\ldots,CC_{r}\}$, $r \geq 1$.
What we intend to show is {\em the direct relation between the stable models of the
completed extended cycles and the stable models of the overall program.}
Indeed, a completed cycle, taken as a program {\it per se,} may or may not have stable
models. 

\begin{theorem}
A program $\Pi$ with decomposition into cycles $\{C_{1},\ldots,C_{r}\}$ 
has stable models only if 
there exists a set of completed extended cycles $\{CC_{1},\ldots,CC_{r}\}$ of
$\Pi$ such that every $CC_{i}$, $i\leq r$, has a stable model. 
\end{theorem}

\ni
For any decomposition of $\Pi$ into completed extended cycles,
we are interested only in those sets $\{S_{1},\ldots,S_{r}\}$ of stable models of, 
respectively,
$\{CC_{1},\ldots,CC_{r}\}$  which agree on shared atoms.
In other words, a consistent set of partial stable models contains one stable
model for each 
of the extended cycles of the decomposition,
and there are no  $S_{i}, S_{j}$ assigning opposite truth values to some atom.

\begin{theorem}
An interpretation $I$ of $\Pi$ is a stable model if and only if
$I = \{S_{1}\cup\ldots\cup S_{r}\}$ where $\{S_{1},\ldots,S_{r}\}$
is a consistent set of stable models for a decomposition
$\{CC_{1},\ldots,CC_{r}\}$ of $\Pi$ into completed extended cycles.
\end{theorem}

Then, from the stable models of the composing cycles,
we are able to obtain the stable models of the program. 
Correspondingly, if we study the conditions for the existence
of stable models of the (extended) cycles, we can find conditions for
the existence of stable models of $\Pi$.
 
It is easy to see that whenever a cycle $C_{\alpha}$ is constrained, then there 
exists a corresponding completed, extended cycle $CC_{\alpha}$ which is a locally 
stratified program; thus, $CC_{\alpha}$ has a unique stable model, which also 
coincide with the Well-founded model.

Assume instead that $\Pi$ contains an unconstrained cycle $C$.
In this case, the unique completed extended cycle associated to $C$ is $C$ itself
({\em trivial} completed extended cycle).
If $C$ is even, then it has the two stable models:

\st $
\begin{array}[l]{l}
M_{C}^{1}=\{a_{i} : i\leq n,\ i=2k+1\}\\
M_{C}^{2}=\{a_{j} : j\leq n,\ j=2k\}
\end{array} $

\st
Vice versa, if $C$ is odd there are no stable models.
In conclusion, we can state the following propositions.

\begin{proposition}
An unconstrained even cycle always has a corresponding (trivial)
completed extended cycle with stable models.
\end{proposition}

\begin{proposition}
An unconstrained odd cycle has no 
corresponding completed extended cycles with a stable model.
\end{proposition}

These considerations allow us to formulate some useful necessary
and sufficient conditions for the existence of stable models. 

In our framework, for instance,
it becomes easy to reformulate the result in \cite{Dun92} saying that every 
call-consistent program has stable models.
Moreover, it is also easy to establish the following.

\begin{proposition}\label{nec1}
A program $\Pi$ has a stable model only if every odd composing
cycle $C$ is constrained.
\end{proposition}

There are situations however, where the odd cycles are 
constrained, but {\em still no stable model exists.}
This happens whenever all possible decompositions of $\Pi$ lead to sets
of partial stable models which are not consistent. I.e.,
there are cycles which require opposite truth values
of some atom, in order to have stable models, e.g.

\st $
\begin{array}[l]{l}
p\ar \no p, \no a.\\
q\ar \no q.\\
q\ar \no a.
\end{array} $

\st
It is possible to identify sufficient conditions for the existence of stable
models, based on ruling out these situations {\sl constructively.}
This is discussed below as well as in our forthcoming work.

\COMMENT
\section{More results on cycles and handles}

\begin{definition}
Two handles are {\em disjoint} if they are composed of rules
belonging to different cycles.
\end{definition}

\begin{definition}
An {\em even handle} is an handle composed only of atoms 
belonging to unconstrained even cycles.
\end{definition}

\begin{definition}
A {\em tightly--even--bounded} program $P$ is a program where,
given the set of odd cycles appearing  in it, we can identify
a corresponding
set of disjoint AND even handles, 
one for each odd cycle. The handles
in this set will be called the {\em crucial handles} of the odd cycles.
\end{definition}

I.e., in a tightly--even--bounded program, not only 
every odd cycle has an even handle, but at least one
even handle for each cycle is disjoint from the even handles
of the other odd cycles. It is important to notice that,
similarly to call--consistency, this condition
is strictly syntactical.

\begin{theorem}
Every tightly--even--bounded program has a stable model.
\end{theorem}

\begin{example}
The following program $P_{t}$ is tightly-even-bounded.
The crucial handles of the odd cycles are indicated in brackets.

\st
\(
EC1\\
a:-\no b.\\
b:-\no a.\)

\st
\(
EC2\\
e:-\no f.\\
f:-\no e.\)

\st
\(
OC1\\
p:-\no p, [\no b].\\
OC2\\
q:-\no q, [\no f].
\)

\st
By combining the partial stable models:

\st
$M_{EC1}^{1} = \{b\}$ \hspace{2cm} $M_{EC2}^{1} = \{f\}$\\
$M_{OC1} = \emptyset$, obtained by assuming $\no b$ false, and\\
$M_{OC2} = \emptyset$, obtained by assuming $\no f$ false,

\stn
we obtain a stable model for the overall program:\\
$M_{P_{t}} = \{b,f\}$, 
which is the unique stable model of $P_{t}$. If we add to the 
program the even cycle:

\st
\(
EC3\\
c:-\no d.\\
d:-\no c.
\)

\st
with partial stable models $M_{EC3}^{1} = \{c\}$ and $M_{EC3}^{2} = \{d\}$,

\stn
then the resulting program $Pt'$ has the two stable models:\\
$M_{Pt'}^{1} = \{b,c,f\}$ \hspace{2cm} $M_{Pt'}^{2} = \{b,d,f\}$
\end{example}

Then, we have identified a class of programs wider than
call--consistent programs, for which a stable model always exists.
The definition however may be relaxed to accommodate
more programs, at the expense of introducing 
some semantical aspects.

\begin{definition}
An {\em even--bounded} program is a program where:
\begin{itemize}
\item
given the set of odd cycles appearing  in it, we can identify
a  set of even handles, one for each cycle. Again,
these handles will be called {\em crucial handles} of the odd cycles.
\item
for each unconstrained even cycle $EC$,  if\\
\(AH\subseteq \{a_{1},\ldots,a_{n}\}\)\\
is the set of atoms belonging to $EC$ that 
occur in the crucial AND handle
of some odd cycle, and \\
\(OH\subseteq \{a_{1},\ldots,a_{n}\}\)\\
is the set of atoms belonging to $EC$ that occur
in the crucial OR handle
of some odd cycle,\\
then either 
\(AH\subseteq M_{EC}^{1}\) and \(OH\subseteq M_{EC}^{2}\) 
or vice versa. We call the candidate stable model
of $EC$ which contains $OH$ (resp. $AH$) $M_{EC}^{OH}$
(resp. $M_{EC}^{AH}$).
\end{itemize}
\end{definition}

The fact that a set of atoms belong to one or the other 
stable model of an even cycle is in itself a semantical notion.
However, 
it can be actually verified syntactically, since, as we have noticed,
the two stable models of an unconstrained even cycle
are simply obtained as a half of the atoms of the cycle.
A tight--even--bounded program is even--bounded.
In even--bounded program however, a crucial 
handle can be an OR handle, and crucial handles are not
required to be disjoint.

\begin{theorem}
Every even--bounded program has a stable model.
\end{theorem}

\begin{example}
The following program $P_{e}$ is even--bounded.\\
\stn
\(
EC1\\
a:-\no b.\\
b:-\no a.\)\\
\stn
\(
OC1\\
p:-\no p, \no a.\)\\
\stn
\(
OC2\\
q:-\no q.\\
q:-\no b.
\)

\st
The literal $\no a$ is an AND even handle for odd cycle OC1 
and the literal $\no b$ is an OR even handle for odd cycle OC2.

\ni
EC1 has the two partial stable models: $M_{EC1}^{1} = \{a\}$ and $M_{EC1}^{2} = \{b\}$,
while the odd cycles have the partial stable models

$M_{OC1} = \emptyset$, by assuming $\no a$ false, and

$M_{OC2} = \{q\}$, by assuming $\no b$ true.

\st
By selecting $M_{EC1}^{1}$, where $\no a$ is false 
and $\no b$ is true, we obtain
the stable model\\
$M_{P_{e}} = \{a,q\}$ 
which is the unique stable model of $P_{e}$. 
\end{example}

It is possible to identify a subclass of even--bounded programs,
composed of programs with a unique stable model,
wider than the class of effectively stratified programs.

\begin{definition}
An even-- bounded program is {\em strongly-- bounded}
if every even cycle is unconstrained,
and has atoms which belong to a
crucial handle of some odd cycle.
\end{definition}

\begin{theorem}
\label{sb}
An even--bounded program which is strongly--bounded has
a unique stable model.
\end{theorem}

For instance, programs $P_{t}$ and $P_{e}$ in the examples
above are strongly bounded (although not
effectively stratified).
Instead, programs $\Pi_1$ and $\Pi_2$ in Example \ref{samegraph1}
are not even--bounded, and still have a (unique) stable model, since 
the odd cycle $OC_1$
has an handle into another odd cycle, $OC_2$, that however
has an handle in the even cycle $EC_1$. $\Pi_3$ instead, cannot have stable models
since $OC_2$ is unconstrained. 

These considerations 
(necessarily informal, for lack of space)
are aimed at suggesting that,
within the proposed formalization,
the above classes may be further enlarged
by means of a finer analysis of the possible interactions
among even and odd cycles.

However, we are now in a position to estimate the maximum number of stable
models  of a program. 
In fact, since in a program which has stable models all handles come 
(directly or indirectly) 
from an even cycle, the number of models is limited by $2^e$ where $e$ is the
number of even cycles.%
 
However, a potential weakness of the approach as 
presented so far, is that it is not easy to identify
cycles, handles and extended cycles. In the next section, we 
propose a new graphical representation of the program, where the above
concepts can be rephrased in an easy way.
\ENDCOMMENT

\section{A new graph representation}\label{newgraph}

In order to reason more directly and more efficiently about cycles and handles,
we introduce a new graph representation of programs, since the usual DG is
not adequate to this aim. On this graph, we should be able of:
detecting by means of efficient algorithms the syntactic features of programs
w.r.t. the classification sketched above; reasoning about the existence
and the number of stable models; computing them. This new graph is similar
to the DG, except it is more accurate for negative dependencies, and thus 
has been called EDG (Extended Dependency Graph).

The definition is based upon distinguishing among rules defining the same atom, 
i.e, having the same head. 
To establish this distinction, we assign to each head an upper index, starting
from $0$, e.g.,
$\{a\ar c,\no b.\ a\ar \no d.\}$ becomes $\{a^0 \ar c^0,\no b^0.\ a^1\ar \no
d^0.\}$.
However, for the sake of clarity, we write $a_i$ instead of $a^{(0)}_i$.
The main idea underlying the next definition is to create, for any atom 
$a$, as many vertices in the graph as the rules with head $a$ (labeled
$a,a^1,a^2$ etc.).

\begin{definition}(Extended dependency graph) (EDG)\label{edg}

\ni
For a logic program $\Pi$, its associated Extended Dependency Graph $EDG(\Pi)$ is the directed 
finite labeled graph $\langle V,E,\{+,-\} \rangle$ defined below.
The main idea underlying the definition of $EDG$ is that of creating, for any atom 
$a$, as many vertices in the graph as the rules with head $a$ (labeled $a,a^1,a^2$ etc.).

\begin{description}
\item[V:1]
For each rule in $\Pi$ there is a vertex $a^{(k)}_i$, where $a_i$ is the name of the head and $k$ is the index of the rule in the definition of $a_i$,


\item[V.2:]
for each atom $u$ never appearing in a head, there is a vertex simply labeled $u$;

\item[E.1:]
for each $c^{(l)}_j\in V$, there is a positive edge $\langle c^{(l)}_j,a^{(k)}_i,+ \rangle$, if and only if $c_j$ appears as a positive condition in
the k-th rule defining $a_i$, and

\item[E.2:]
for each $c^{(l)}_j\in V$, there is a negative edge $\langle c^{(l)}_j,a^{(k)}_i,-
\rangle$, if and only if $c_j$ appears as a negative condition in the k-th rule defining $a_i$.
\end{description}
\end{definition}

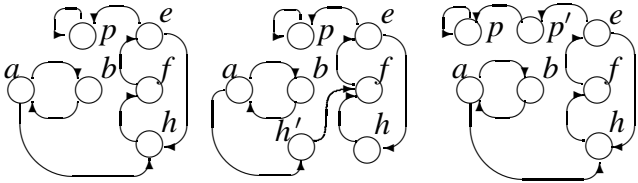
\begin{figure}[htbp]
\setlength{\unitlength}{2000sp} 
\begingroup\makeatletter\ifx\SetFigFont\undefined
\def\x#1#2#3#4#5#6#7\relax{\def\x{#1#2#3#4#5#6}}%
\expandafter\x\fmtname xxxxxx\relax \def\y{splain}%
\ifx\x\y   
\gdef\SetFigFont#1#2#3{%
  \ifnum #1<17\tiny\else \ifnum #1<20\small\else
  \ifnum #1<24\normalsize\else \ifnum #1<29\large\else
  \ifnum #1<34\Large\else \ifnum #1<41\LARGE\else
     \huge\fi\fi\fi\fi\fi\fi
  \csname #3\endcsname}%
\else
\gdef\SetFigFont#1#2#3{\begingroup
  \count@#1\relax \ifnum 25<\count@\count@25\fi
  \def\x{\endgroup\@setsize\SetFigFont{#2pt}}%
  \expandafter\x
    \csname \romannumeral\the\count@ pt\expandafter\endcsname
    \csname @\romannumeral\the\count@ pt\endcsname
  \csname #3\endcsname}%
\fi
\fi\endgroup
\begin{picture}(7824,2197)(106,-1465)
\thinlines
\put(457,-486){\vector( 0, 1){0}}
\put(738,-486){\oval(562,420)[bl]}
\put(738,-486){\oval(560,420)[br]}
\put(1018,-205){\vector( 0,-1){0}}
\put(738,-205){\oval(560,420)[tr]}
\put(738,-205){\oval(562,420)[tl]}
\put(3157,-486){\vector( 0, 1){0}}
\put(3438,-486){\oval(562,420)[bl]}
\put(3438,-486){\oval(560,420)[br]}
\put(3718,-205){\vector( 0,-1){0}}
\put(3438,-205){\oval(560,420)[tr]}
\put(3438,-205){\oval(562,420)[tl]}
\put(6560,-193){\vector( 0,-1){0}}
\put(6278,-193){\oval(564,422)[tr]}
\put(6278,-193){\oval(564,422)[tl]}
\put(5996,-476){\vector( 0, 1){0}}
\put(6278,-476){\oval(564,422)[bl]}
\put(6278,-476){\oval(564,422)[br]}
\put(1756,293){\vector( 1, 0){0}}
\put(1756, 13){\oval(422,560)[tl]}
\put(1756, 13){\oval(422,562)[bl]}
\put(1756,-418){\vector( 1, 0){0}}
\put(1756,-699){\oval(422,562)[tl]}
\put(1756,-699){\oval(422,560)[bl]}
\put(7293,-406){\vector( 1, 0){0}}
\put(7293,-689){\oval(424,566)[tl]}
\put(7293,-689){\oval(424,564)[bl]}
\put(7278,300){\vector( 1, 0){0}}
\put(7278, 18){\oval(424,564)[tl]}
\put(7278, 18){\oval(424,566)[bl]}
\put(1502,482){\oval(560,420)[tr]}
\put(1502,482){\oval(562,420)[tl]}
\put(1221,482){\vector( 0,-1){0}}
\put(4202,466){\oval(560,422)[tr]}
\put(4202,466){\oval(562,422)[tl]}
\put(3921,466){\vector( 0,-1){0}}
\put(7051,510){\oval(564,424)[tr]}
\put(7051,510){\oval(566,424)[tl]}
\put(6768,510){\vector( 0,-1){0}}
\put(6241,495){\oval(564,424)[tr]}
\put(6241,495){\oval(564,424)[tl]}
\put(5959,495){\vector( 0,-1){0}}
\put(5717,562){\oval(324,240)[tr]}
\put(5717,512){\oval(340,340)[tl]}
\put(5667,512){\oval(240,324)[bl]}
\put(5667,350){\vector( 1, 0){0}}
\put(3615,532){\oval(322,238)[tr]}
\put(3615,482){\oval(338,338)[tl]}
\put(3566,482){\oval(240,322)[bl]}
\put(3566,321){\vector( 1, 0){0}}
\put(900,547){\oval(322,238)[tr]}
\put(900,497){\oval(338,338)[tl]}
\put(851,497){\oval(240,322)[bl]}
\put(851,336){\vector( 1, 0){0}}
\put(4472,-423){\vector( 1, 0){0}}
\put(4472,-704){\oval(422,562)[tl]}
\put(4472,-704){\oval(422,560)[bl]}
\put(4442,283){\vector( 1, 0){0}}
\put(4442,  3){\oval(422,560)[tl]}
\put(4442,  3){\oval(422,562)[bl]}
\put(7638,-1049){\vector(-1, 0){0}}
\put(7638,-344){\oval(566,1410)[br]}
\put(7638,-344){\oval(566,1412)[tr]}
\put(6872,-508){\oval(1982,1896)[bl]}
\put(6872,-1243){\oval(1228,426)[br]}
\put(7486,-1243){\vector( 0, 1){0}}
\put(1291,-508){\oval(1980,1806)[bl]}
\put(1291,-1198){\oval(1230,426)[br]}
\put(1906,-1198){\vector( 0, 1){0}}
\put(2836,-756){\oval(300,826)[tl]}
\put(3329,-756){\oval(1286,1286)[bl]}
\put(3329,-1213){\oval(904,372)[br]}
\put(3781,-1213){\vector( 0, 1){0}}
\put(2091,-1048){\vector(-1, 0){0}}
\put(2091,-347){\oval(560,1402)[br]}
\put(2091,-347){\oval(560,1404)[tr]}
\put(4806,-1063){\vector(-1, 0){0}}
\put(4806,-362){\oval(560,1402)[br]}
\put(4806,-362){\oval(560,1404)[tr]}
\put(3788,-1047){\circle{320}}
\put(4630,-345){\circle{320}}
\put(316,-345){\circle{320}}
\put(7478,-1040){\circle{320}}
\put(5855,371){\circle{320}}
\put(1156,-343){\circle{320}}
\put(1081,332){\circle{320}}
\put(1906,-1018){\circle{320}}
\put(1906,-343){\circle{320}}
\put(1906,332){\circle{320}}
\put(3024,-340){\circle{320}}
\put(3781,-343){\circle{320}}
\put(4606,-1093){\circle{320}}
\put(3781,332){\circle{320}}
\put(7456,-343){\circle{320}}
\put(6631,-343){\circle{320}}
\put(5881,-343){\circle{320}}
\put(6631,407){\circle{320}}
\put(4621,332){\circle{320}}
\put(7456,362){\circle{320}}
\put(4399,-328){\vector( 1, 0){ 42}}
\multiput(4399,-331)(-11.35135,-1.89189){3}{\makebox(1.6667,11.6667){\SetFigFont{5}{6}{rm}.}}
\multiput(4376,-333)(-9.24323,-1.54054){4}{\makebox(1.6667,11.6667){\SetFigFont{5}{6}{rm}.}}
\multiput(4348,-336)(-9.62163,-1.60361){4}{\makebox(1.6667,11.6667){\SetFigFont{5}{6}{rm}.}}
\multiput(4319,-340)(-10.32433,-1.72072){4}{\makebox(1.6667,11.6667){\SetFigFont{5}{6}{rm}.}}
\multiput(4288,-345)(-9.67567,-1.61261){4}{\makebox(1.6667,11.6667){\SetFigFont{5}{6}{rm}.}}
\multiput(4259,-350)(-9.33333,-2.33333){4}{\makebox(1.6667,11.6667){\SetFigFont{5}{6}{rm}.}}
\multiput(4231,-357)(-8.00000,-2.66667){4}{\makebox(1.6667,11.6667){\SetFigFont{5}{6}{rm}.}}
\multiput(4207,-365)(-11.03450,-4.41380){3}{\makebox(1.6667,11.6667){\SetFigFont{5}{6}{rm}.}}
\multiput(4185,-374)(-9.00000,-6.00000){3}{\makebox(1.6667,11.6667){\SetFigFont{5}{6}{rm}.}}
\multiput(4167,-386)(-7.62295,-6.35246){3}{\makebox(1.6667,11.6667){\SetFigFont{5}{6}{rm}.}}
\multiput(4152,-399)(-6.42000,-8.56000){3}{\makebox(1.6667,11.6667){\SetFigFont{5}{6}{rm}.}}
\multiput(4139,-416)(-8.29410,-13.82350){2}{\makebox(1.6667,11.6667){\SetFigFont{5}{6}{rm}.}}
\multiput(4131,-430)(-3.24140,-8.10350){3}{\makebox(1.6667,11.6667){\SetFigFont{5}{6}{rm}.}}
\multiput(4124,-446)(-3.00000,-9.00000){3}{\makebox(1.6667,11.6667){\SetFigFont{5}{6}{rm}.}}
\multiput(4118,-464)(-2.38235,-9.52940){3}{\makebox(1.6667,11.6667){\SetFigFont{5}{6}{rm}.}}
\multiput(4113,-483)(-1.83785,-11.02710){3}{\makebox(1.6667,11.6667){\SetFigFont{5}{6}{rm}.}}
\multiput(4109,-505)(-1.91890,-11.51340){3}{\makebox(1.6667,11.6667){\SetFigFont{5}{6}{rm}.}}
\put(4105,-528){\line( 0,-1){ 25}}
\multiput(4103,-553)(-1.43243,-8.59460){4}{\makebox(1.6667,11.6667){\SetFigFont{5}{6}{rm}.}}
\put(4100,-579){\line( 0,-1){ 27}}
\put(4099,-606){\line( 0,-1){ 28}}
\put(4098,-634){\line( 0,-1){ 28}}
\put(4096,-662){\line( 0,-1){ 28}}
\put(4095,-690){\line( 0,-1){ 28}}
\put(4094,-718){\line( 0,-1){ 27}}
\put(4093,-745){\line( 0,-1){ 26}}
\put(4092,-771){\line( 0,-1){ 25}}
\multiput(4090,-796)(-1.89190,-11.35140){3}{\makebox(1.6667,11.6667){\SetFigFont{5}{6}{rm}.}}
\multiput(4088,-819)(-1.82430,-10.94580){3}{\makebox(1.6667,11.6667){\SetFigFont{5}{6}{rm}.}}
\multiput(4085,-841)(-1.58110,-9.48660){3}{\makebox(1.6667,11.6667){\SetFigFont{5}{6}{rm}.}}
\multiput(4082,-860)(-1.80770,-9.03850){3}{\makebox(1.6667,11.6667){\SetFigFont{5}{6}{rm}.}}
\multiput(4078,-878)(-3.76470,-15.05880){2}{\makebox(1.6667,11.6667){\SetFigFont{5}{6}{rm}.}}
\multiput(4074,-893)(-4.70000,-14.10000){2}{\makebox(1.6667,11.6667){\SetFigFont{5}{6}{rm}.}}
\multiput(4069,-907)(-3.65515,-9.13787){3}{\makebox(1.6667,11.6667){\SetFigFont{5}{6}{rm}.}}
\multiput(4061,-925)(-9.23080,-13.84620){2}{\makebox(1.6667,11.6667){\SetFigFont{5}{6}{rm}.}}
\multiput(4052,-939)(-11.00000,-11.00000){2}{\makebox(1.6667,11.6667){\SetFigFont{5}{6}{rm}.}}
\multiput(4041,-950)(-12.00000,-8.00000){2}{\makebox(1.6667,11.6667){\SetFigFont{5}{6}{rm}.}}
\multiput(4029,-958)(-14.00000,-7.00000){2}{\makebox(1.6667,11.6667){\SetFigFont{5}{6}{rm}.}}
\multiput(4015,-965)(-15.05880,-3.76470){2}{\makebox(1.6667,11.6667){\SetFigFont{5}{6}{rm}.}}
\multiput(4000,-969)(-15.96150,-3.19230){2}{\makebox(1.6667,11.6667){\SetFigFont{5}{6}{rm}.}}
\multiput(3984,-972)(-15.00000,-3.00000){2}{\makebox(1.6667,11.6667){\SetFigFont{5}{6}{rm}.}}
\put(3969,-975){\line(-1, 0){ 14}}
\multiput(3955,-976)(-10.86490,-1.81082){2}{\makebox(1.6667,11.6667){\SetFigFont{5}{6}{rm}.}}
\put(2056,-868){\makebox(0,0)[lb]{\smash{\SetFigFont{12}{14.4}{it}h}}}
\put(2056,-193){\makebox(0,0)[lb]{\smash{\SetFigFont{12}{14.4}{it}f}}}
\put(2806,-193){\makebox(0,0)[lb]{\smash{\SetFigFont{12}{14.4}{it}a}}}
\put(3931,-193){\makebox(0,0)[lb]{\smash{\SetFigFont{12}{14.4}{it}b}}}
\put(4756,-193){\makebox(0,0)[lb]{\smash{\SetFigFont{12}{14.4}{it}f}}}
\put(1306,-193){\makebox(0,0)[lb]{\smash{\SetFigFont{12}{14.4}{it}b}}}
\put(7606,482){\makebox(0,0)[lb]{\smash{\SetFigFont{12}{14.4}{it}e}}}
\put(7606,-193){\makebox(0,0)[lb]{\smash{\SetFigFont{12}{14.4}{it}f}}}
\put(7606,-868){\makebox(0,0)[lb]{\smash{\SetFigFont{12}{14.4}{it}h}}}
\put(6781,-193){\makebox(0,0)[lb]{\smash{\SetFigFont{12}{14.4}{it}b}}}
\put(5656,-193){\makebox(0,0)[lb]{\smash{\SetFigFont{12}{14.4}{it}a}}}
\put(6106,332){\makebox(0,0)[lb]{\smash{\SetFigFont{12}{14.4}{it}p}}}
\put(6856,332){\makebox(0,0)[lb]{\smash{\SetFigFont{12}{14.4}{it}p$^{{\small \prime}}$ }}}
\put(106,-193){\makebox(0,0)[lb]{\smash{\SetFigFont{12}{14.4}{it}a}}}
\put(2026,497){\makebox(0,0)[lb]{\smash{\SetFigFont{12}{14.4}{it}e}}}
\put(1306,287){\makebox(0,0)[lb]{\smash{\SetFigFont{12}{14.4}{it}p}}}
\put(4006,287){\makebox(0,0)[lb]{\smash{\SetFigFont{12}{14.4}{it}p}}}
\put(4771,482){\makebox(0,0)[lb]{\smash{\SetFigFont{12}{14.4}{it}e}}}
\put(3466,-973){\makebox(0,0)[lb]{\smash{\SetFigFont{12}{14.4}{it}h$^{{\small
\, \prime}}$ }}}
\put(4696,-883){\makebox(0,0)[lb]{\smash{\SetFigFont{12}{14.4}{it}h}}}
\end{picture}
\caption{$EDG(\Pi_3)$ (left), $EDG(\Pi_1)$ (center)
and $EDG(\Pi_2)$ (right). }\label{fig:edgs}

Notice that both $DG(\Pi_1),DG(\Pi_2)$ and $DG(\Pi_3)$ correspond to $EDG(\Pi_3)$.
\label{edgsfig}\end{figure}

The definition of $EDG$ extends that of DG in the sense that for programs 
where atoms are defined by at most one rule the two coincide.
Consider in Figure \ref{edgsfig} the $EDGs$ of the programs in Example \ref{samegraph1}. 
As all conditions in $\Pi_1$, $\Pi_2$ and $\Pi_3$ are negative, for the sake of
simplicity, the `-' labels are omitted from edges.

The main idea underlying the definition of $EDG$ is that of creating, for any atom 
$a$, as many vertices in the graph as the rules with head $a$ (labeled $a,a^1,a^2$ etc.).
For instance, in $EDG(\Pi_1)$ (center of Figure \ref{edgsfig}) arc
$\langle h,f,-\rangle$ represents rule $\{f\ar \no h.\}$.
On the graph, we clearly see the cycles, and also the handles.
In fact, rule $\{f\ar \no h.\}$ {\em must} be represented by the two arcs 
$\langle h,f,-\rangle$ and $\langle h^\prime,f,-\rangle$ since truth of $h$ may
depend on any of its defining rules; the second one is auxiliary to the cycle,
and corresponds to an OR handle. 
Therefore, the cycle has
an OR handle {\em if and only if  there is an incoming arc originated in a duplication of one of the atoms of the cycle}. 
In this case, the arc $\langle h^\prime,f,-\rangle$ represents the $OR$ handle of
$OC_2$.
In the same graph, arc $\langle e,p,-\rangle$ represents instead the $AND$ handle
of $OC_1$.
Therefore, a cycle has an AND handle {\em if and only if} there exists an incoming
arc into that cycle in the $EDG$, originated in (any duplication of) 
an atom not belonging to the cycle itself. {\em A cycle with no
incoming arcs is unconstrained}.

\ni
It is easy to see that the $EDG$ of a program is isomorphic to the program itself.
Consequently, the $EDG$ conveys enough information for reasoning about 
stable models of the program.

\section{Coloring {\it EDGs}}
This section describes how the $EDG$ can be used to study
the stable models in terms of graph coloring.
Let us define a {\it coloring} as an assignment of nodes of a graph to colors, 
e.g. $\nu: V \rar \{green,\ red\}$. 
An interpretation corresponds to a coloring, where all the true atoms are green,
and all the others are red.

We now specify which colorings we intend to rule out, since they trivially
correspond to inconsistencies.


\begin{definition} (non-admissible coloring)

\ni
A coloring $\nu: V \rar \{green,\ red\}$ is non-admissible for $\langle V,E
\rangle=EDG(\Pi)$ if and only if

\begin{enumerate}

\item $\exists i.\nu(v_i)=green$ and $\exists j. (v_i,v_j,-) \in E$ 
and $\nu(v_j)=green$, or

\item $\exists i.\nu(v_i)=red$ and $\forall j. (v_j,v_i,-) \in E$ and $\nu(v_j)=red$.

\end{enumerate}

\ni
To sum it up, {\bf green nodes cannot be adjacent and edges to 
a red node cannot all come from red nodes.}

\end{definition}

A coloring for $EDG(\Pi)$ is admissible unless it is not admissible. 
A partial coloring is
admissible if all its {\sl completions (intuitively)} are.

\begin{example} What are the admissible colorings for 
$EDG(\Pi_1)$ in Example \ref{samegraph1}?

\begin{figure}[htbp]
\input{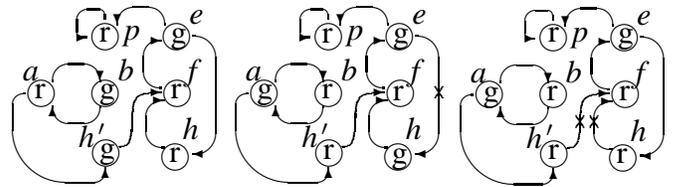}
\caption{An admissible coloring of $EDG(\Pi_1)$ (on the left) and two
not admissible ones (center and right, resp.) of $EDG(\Pi_1)$ with g=green, r=red and
X=admissibility violation.}
\label{fig:edgs1}
\end{figure}

In the center coloring above, arc $\langle e,h,-\rangle$ violates admissibility. 
In fact, it corresponds to rule $h \ar \no e$ in $\Pi_1$. 
If $e$ is true/green, then by the rule $h$ cannot be concluded true/green.
As a matter of fact, both $e$ and $h$ are true in the stable model of $\Pi_1$, 
but the truth of $h$ comes from, intuitively, labeling $h^\prime$ green in the
first coloring.
In the right coloring above, admissibility is violated by arcs
$\langle h^\prime,f,-\rangle$ and $\langle h,f,-\rangle$ which, together, represent the 
rule $f\ar \no h$ of $\Pi_1$. 
When all $h$s are red, we conclude $h$ false and --by the above rule--, $f$ true/green.
\end{example}

\ni
Now, we are able to define a notion of admissible coloring for EDG's of Kernel
programs.

\begin{theorem}
An interpretation $\cali$ is a stable model of $\Pi$ if and only if 
it corresponds to
an admissible coloring of $EDG(\Pi)$.
\end{theorem}

We are implementing a practical system that computes stable models on the EDG \cite{BCP99}.
The coloring procedure is, experimentally, very sensitive to the choice
of heuristic methods for starting the coloring itself from ``relevant'' nodes. 
In fact, presently the choice of the starting nodes is guided by 
the concept of extended cycle described earlier: we try identify
nodes corresponding to crucial handles, and start from them. 
A main topic for our research now is clearly the study of new heuristic methods,
as well as adapting existing solutions from graph theory.

\section*{Acknowledgments}
Thanks to Chitta Baral and Michael Gelfond for constant encouragement
 in the pursuit of this research.

\bibliographystyle{aaai}

\begin{thebibliography}{SA95}

\bibitem[Apt \& Bol 1994]{AptBol94}
Apt, K. R. and Bol, R., 1994.
\newblock {\it Logic programming and negation: a survey,}
\newblock J. of Logic Programming, 19/20.

\bibitem[Baral & Gelfond 1994]{BarGel94}
Baral, C. and Gelfond. M., 1994.
\newblock {\it Logic programming and knowledge representation,}
\newblock J. of Logic Programming, 19/20.

\bibitem[Brignoli et al. 1999]{BCP99}
Brignoli G., Costantini S. and Provetti A., 1999.
\newblock {\it A Graph Coloring algorithm for stable models generation.} 
\newblock Univ. of Milan Technical Report, submitted for publication.

\bibitem[Costantini, 1995]{Cos95}
Costantini S., 1995.
\newblock {\it Contributions to the stable model semantics of logic programs with negation,}
\newblock Theoretical Computer Science, 149.

\bibitem[Cholewin\'nski et al. 1996]{ChoMarTru96}
Cholewi\'nski P., Marek W. and Truszczy\'nski M., 1996.
\newblock {\it Default reasoning system DeReS.}
\newblock Proc. of KR96, Morgan-Kauffman, pp. 518-528.

\bibitem[Cholewi\'nski \& Truszczy\'nski 1996]{ChoTru96}
Cholewi\'nski P. and Truszczy\'nski M., 1996.
\newblock {\it Extremal problems in logic programming and stable model computation.}
\newblock Proc. of IJCSLP'96, pp. 408--422.
\newblock Also in J. of Logic Programming, 38(1999), pp. 219--242.

\bibitem[Dimopoulos 1996]{Dim96}
Dimopoulos Y., 1996.
\newblock {\it On Computing Logic Programs,}
\newblock J. of Automated Reasoning, 17:259--289.

\bibitem[Dung 1992]{Dun92}
Dung P.M., 1992.
\newblock {\it On the Relation between Stable and 
Well--Founded Semantics of Logic Programs,}
\newblock Theoretical Computer Science, 105.

\bibitem[Dunng \& Kanchanasut 1989]{Dun89}
Dung P.M. and Kanchanasut, 1989.
\newblock {\it Logic programming and stable model computation,}
\newblock Proc. of NACLP'89.

\bibitem[Eiter et al. 1997]{Eit+97}
Eiter, T., Leone, N., Mateis, C., Pfeifer, G., and Scarcello, F., 1997.
\newblock {\it A deductive system for non-monotonic reasoning.} 
\newblock Proc. Of the 4 th LPNMR Conference, Springer Verlag, 
LNCS 1265, pp. 363--374.

\bibitem[Gelfond \& Lifschitz 1988]{GelLif88}
Gelfond, M.  and Lifschitz,  V., 1988.
\newblock {\it The stable model semantics for
logic programming,} Proc. of 5th ILPS conference, pp. 1070--1080.

\bibitem[Marek \& Truszczy\'nski 1991]{MarTru91}
Marek, W., and Truszczy\'nski M., 1991.
\newblock {\it Autoepistemic Logic.}
\newblock The Journal of the ACM,38:588--619.

\bibitem[Marek \& Truszczy\'nski 1999]{MarTru99}
Marek, W., and Truszczy\'nski M., 1999.
\newblock {\it Stable models and an alternative logic programming paradigm.}
\newblock The Journal of Logic Programming.

\bibitem[Niemel\"a \& Simons 1998]{NieSim98}
Niemel\"a I. and Simons P., 1998.
\newblock {\it Logic programs with stable model semantics 
as a constraint programming paradigm.} 
\newblock Proc. of NM'98 workshop. Extended version submitted for publication.

\bibitem[Sacc\`a \& Zaniolo 1997]{SacZan97}
Sacc\`a D. and Zaniolo C., 1997.
\newblock {\it Deterministic and Non-Deterministic Stable Models.}
\newblock J. of Logic and Computation.

\bibitem[Simons 1997]{Sim97}
Simons P., 1997.
\newblock {\it Towards Constraint Satisfaction through Logic Programs and the 
Stable Models Semantics,} Helsinki Univ. of Technology R.R. A:47.

\bibitem[Subrahmanian et al. 1995]{SubNauVag95}
Subrahmanian, V.S., Nau D., and Vago C., 1995.
\newblock {\it WFS + branch and bound = stable models,} 
IEEE Trans. on Knowledge and Data Engineering, 7(3):362--377.

\bibitem[Van Gelder et al. 1990]{wfm}
Van Gelder A., Ross K.A. and Schlipf J., 1990.
\newblock {\it The Well-Founded Semantics for General Logic Programs}.
\newblock  Journal of the ACM Vol. 38 N. 3.

\bibitem[Chen \& Warren 1996]{CheWar96}
Chen W., and Warren D.S., 1996.
\newblock {\it Computation of stable models and its integration with logical
  query processing,} IEEE Trans. on Data and Knowledge Engineering, 8(5):742--747.

\end{thebibliography}

\end{document}